\documentclass[letterpaper]{article} 
\usepackage{aaai2026}  
\usepackage{times}  
\usepackage{helvet}  
\usepackage{courier}  
\usepackage[hyphens]{url}  
\usepackage{graphicx} 
\usepackage{natbib}  
\usepackage{caption} 
\usepackage{algorithm}
\usepackage{algorithmic}
\usepackage{amsmath}
\usepackage{amssymb}
\usepackage{multirow}
\usepackage{etoolbox}
\usepackage{booktabs}
\usepackage{xcolor}

\usepackage{newfloat}
\usepackage{listings}
\DeclareCaptionStyle{ruled}{labelfont=normalfont,labelsep=colon,strut=off} 
\floatstyle{ruled}
\newfloat{listing}{tb}{lst}{}
\floatname{listing}{Listing}
\title{Dependency Structure Augmented Contextual Scoping Framework for Multimodal Aspect-Based Sentiment Analysis}
\author{
    Hao Liu\textsuperscript{\rm 1},
    Lijun He\textsuperscript{\rm 1},
    Jiaxi Liang\textsuperscript{\rm 1},
    Zhihan Ren\textsuperscript{\rm 1},
    Haixia Bi\textsuperscript{\rm 1},
    Fan Li\textsuperscript{\rm 1},
}
\affiliations{
    \textsuperscript{\rm 1}Xi'an Jiaotong University, Xi'an, 710049, China\\
    haoliu88@stu.xjtu.edu.cn,
    lijunhe@mail.xjtu.edu.cn,
    liangjiaxi@stu.xjtu.edu.cn,
    renzh@stu.xjtu.edu.cn,
    haixia.bi@xjtu.edu.cn,
    lifan@mail.xjtu.edu.cn
}

\usepackage{bibentry}

\begin{document}

\maketitle

\begin{abstract}
Multimodal Aspect-Based Sentiment Analysis (MABSA) seeks to extract fine-grained information from image-text pairs to identify aspect terms and determine their sentiment polarity. However, existing approaches often struggle to simultaneously address three core challenges: Sentiment Cue Perception (SCP), Multimodal Information Misalignment (MIM), and Semantic Noise Elimination (SNE).
To overcome these limitations, we propose DASCO (\textbf{D}ependency Structure \textbf{A}ugmented \textbf{Sco}ping Framework), a fine-grained scope-oriented framework. We introduce a \textit{continue pretraining strategy}, combining aspect-oriented enhancement, image-text matching, and aspect-level sentiment-sensitive cognition, which strengthens the model's perception of aspect terms and sentiment cues while achieving effective image-text alignment, addressing key challenges like SCP and MIM.
Furthermore, DASCO integrates a \textit{syntactic-semantic} dual-branch architecture that leverages dependency structures to construct \textit{target-specific scopes} and employs \textit{adaptive scope interaction} to guide the model to focus on target-relevant context while filtering out noise, thereby alleviating the SNE challenge. 
Extensive experiments on two benchmark datasets across three subtasks demonstrate that DASCO achieves state-of-the-art performance, with notable gains in JMASA (+2.3\% F1 and +3.5\% precision on Twitter2015). The source code is available at \url{https://github.com/LHaoooo/DASCO}.
\end{abstract}

\section{Introduction}

Multimodal Aspect-Based Sentiment Analysis (MABSA) \cite{DBLP:conf/emnlp/JuZXLLZZ21, DBLP:conf/acl/LingYX22, DBLP:conf/acl/ZhouGLYZY23, DBLP:conf/aaai/PengLWZZ24} is a cutting-edge research direction in affective computing, focusing on extracting fine-grained associative semantic and sentiment information from heterogeneous modal data (text-image pairs). Unlike traditional unimodal sentiment analysis approaches, which relies solely on textual data, MABSA leverages cross-modal feature alignment to enhance the identification of aspect terms and their sentiment orientations.
The MABSA framework encompasses three key subtasks: Multimodal Aspect Term Extraction (MATE), Multimodal Aspect Sentiment Classification (MASC), and Joint Multimodal Aspect Sentiment Analysis (JMASA).
Specially, MATE \cite{DBLP:conf/acl/LiLSWZW024, DBLP:conf/mm/LiYYWYX24, DBLP:conf/coling/HuangXLYL24} concentrates on identifying all aspect terms in sentences; MASC \cite{DBLP:conf/coling/FengL0G24, DBLP:conf/ijcai/Yu019} aims to determine the sentiment polarity (positive, negative, or neutral) for each identified aspect term; and JMASA \cite{DBLP:conf/emnlp/JuZXLLZZ21, DBLP:conf/acl/LingYX22, DBLP:conf/acl/ZhouGLYZY23} jointly models aspect-sentiment pairs in an end-to-end fashion.

\begin{figure}[t]
    \centering
    \includegraphics[width=1\linewidth]{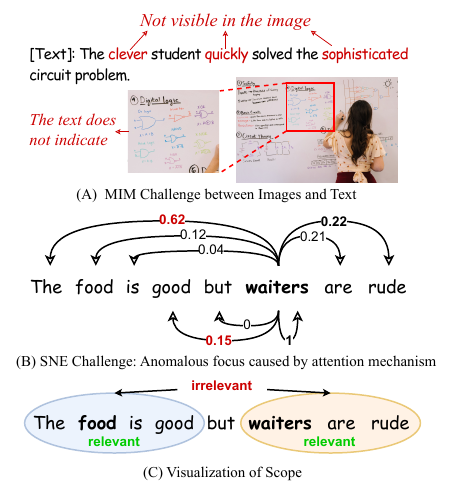}
    \caption{Partial challenges in MABSA and Visualization of Scope.}
    \label{fig:dasco1}
\end{figure}

Existing research on MABSA \cite{DBLP:conf/emnlp/JuZXLLZZ21, DBLP:journals/ipm/YangNY22, DBLP:conf/coling/LiuLYZW25, DBLP:journals/taffco/YuCX23, DBLP:journals/inffus/XiaoWXLJH24, DBLP:conf/ijcai/ZhuSGY024, DBLP:conf/icmcs/LiuHL24, DBLP:journals/tkde/MuNWXZL24, DBLP:conf/icmcs/LiuZLZSZH24, DBLP:conf/emnlp/Xiao0ZHC24} mostly relies on using pre-trained models (such as BERT, RoBERTa, ResNet, CLIP \cite{DBLP:conf/naacl/DevlinCLT19, DBLP:journals/corr/abs-1907-11692, DBLP:conf/icml/RadfordKHRGASAM21, DBLP:conf/iclr/DosovitskiyB0WZ21}) to extract visual and textual features. These pre-trained models usually use general vision-language understanding tasks (e.g., next sentence prediction, masked language modeling, image classification, and image-text matching) as their pre-training objectives. However, these objectives differ significantly from MABSA's goals \cite{DBLP:conf/acl/LingYX22}, resulting in misalignment when capturing fine-grained aspects, opinions, and sentiment cues across modalities. We summarize this gap as the \textit{Sentiment Cue Perception (SCP)} challenge.

Furthermore, the inherent heterogeneity of image-text data naturally leads to information asymmetry and mismatches, complicating multimodal feature alignment and fusion. As shown in Figure \ref{fig:dasco1}.A, elements like ``Digital logic'' appear only in the image, while ``circuit problem'' in the text loosely relates to the image content. Additionally, sentiment-rich terms such as ``clever'' or ``sophisticated'' are also hard to infer visually. We define this phenomenon as the \textit{Multimodal Information Misalignment (MIM)} challenge.

To address MIM, previous work \cite{DBLP:conf/emnlp/JuZXLLZZ21, DBLP:journals/ipm/YangNY22, DBLP:conf/icmcs/LiuHL24, DBLP:conf/coling/LiuLYZW25} have explored text-guided image alignment and cross-modal relationship modeling, typically using attention mechanisms. However, when handling multi-aspect inputs, attention can propagate noise and weaken key semantics. As shown in Figure \ref{fig:dasco1}.B, in the sentence ``The food is good but waiters are rude'', BERT \cite{DBLP:conf/naacl/DevlinCLT19} assigns disproportionate attention to irrelevant tokens, focusing more on ``The'' (0.62) than on the core sentiment word ``rude'' (0.22) and even maintains abnormal attention to the positive sentiment word ``good'' (0.15). This counterintuitive pattern of attention distribution is attributed to noise interference in the attention mechanism, which we define as the \textit{Semantic Noise Elimination (SNE)} challenge.

To tackle the challenges mentioned above, we propose a Dependency Structure Augmented Contextual Scoping Framework (DASCO), which operates in two phases. In the first phase, we introduce aspect-oriented enhancement (AOE) and aspect-level sentiment-sensitive cognition (ASSC) pre-training tasks for the base models (Qformer \cite{DBLP:conf/icml/0008LSH23} and text encoder). Unlike conventional objectives, our design explicitly focuses on aspect and sentiment cues, making them better suited for MABSA and effectively alleviating the SCP challenge.
To bridge modality gaps, we construct an image Scene Graph dataset using \verb|GPT-4o|. These graphs offer structured, text-based descriptions of visual content, enriching textual input and facilitating feature alignment. The Scene Graph is encoded alongside the image through Qformer, enabling cross-modal mapping into a unified textual space.
We also retain the standard image-text matching (ITM) pre-training task from conventional MABSA designs \cite{DBLP:conf/acl/LingYX22, DBLP:conf/aaai/PengLWZZ24}. Together, the Scene Graph and the ITM task effectively mitigate the MIM problem.

In the second phase, DASCO integrates a syntactic-semantic dual-branch architecture with an adaptive scope interaction mechanism to enhance context modeling. Although dependency structures have proven effective in text-only ABSA \cite{DBLP:conf/coling/ChenTS20, DBLP:conf/acl/MaHLYLYW23, DBLP:conf/naacl/DaiYSLQ21, DBLP:conf/acl/TangJLZ20}, they remain underexplored in MABSA. 
DASCO leverages dependency trees to define target-specific scopes (Figure \ref{fig:dasco1}.C) and applies adaptive scope interaction to perform semantic interactions across scopes and across graphs, guiding the model to focus on semantics within the target-specific scope. This effectively filters out irrelevant noise and alleviates the SNE challenge.

In summary, our contributions are:
\begin{itemize}
    \item We propose DASCO, a two-phase framework achieving state-of-the-art performance on three MABSA subtasks.
    \item To address \textit{SCP} and \textit{MIM}, we design aspect- and sentiment-aware continue pretraining tasks and construct a Scene Graph dataset to enhance multimodal alignment and cue perception.
    \item To tackle \textit{SNE}, we introduce a syntactic-semantic dual-branch architecture equipped with an adaptive scope interaction mechanism, which models target-specific scopes to enable precise aspect and sentiment reasoning while effectively filtering irrelevant noise.
\end{itemize}

\section{Methodology}
\subsection{Preliminaries}

\subsubsection{Scene Graph Dataset Construction}
\label{secsg}
\begin{figure}[t]
    \centering
    \includegraphics[width=1\linewidth]{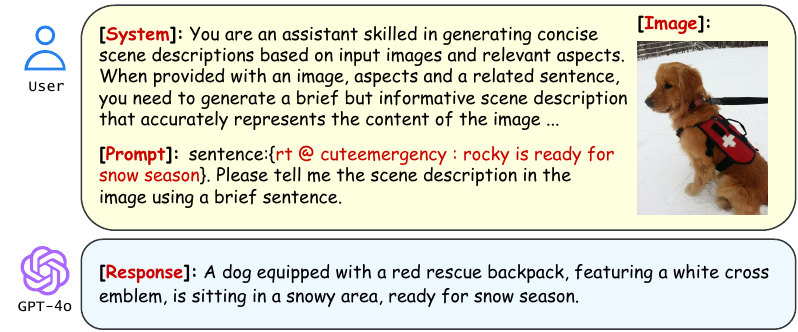}
    \caption{System instructions and prompts for scene graph generation using GPT-4o.}
    \label{fig:dasco2}
\end{figure}

Inspired by Atlantis \cite{DBLP:journals/inffus/XiaoWXLJH24}, we incorporate \textit{scene graph} to enhance aspect-specific image semantic extraction. While we adopt Qformer \cite{DBLP:conf/icml/0008LSH23} for visual-textual bridging, the original Qformer structure suffers from spatial positional information loss \cite{DBLP:journals/corr/abs-2405-20985}. To overcome this limitation, we utilize \verb|GPT-4o| to generate structured textual descriptions of images (``Scene Graphs''). This approach not only improves cross-modal alignment but also preserves spatial semantics through Qformer's language modeling capabilities. We implement this solution using carefully designed system-level instructions and prompts for \verb|GPT-4o|. A complete example is given in the \textit{Supplementary Materials}, with a short snippet as seen in Figure \ref{fig:dasco2}.

\subsubsection{Pretraining Dataset Construction}
To support continue pretraining, we construct derived training samples from the original annotated data, each comprising a text $T$, image $V$, scene graph $S$, aspect term(s) $A$, and sentiment polarity $y_c$ of aspect term(s).
We construct the following three sub-datasets for pretraining tasks:

\begin{figure*}[t]
    \centering
    \includegraphics[width=1\linewidth]{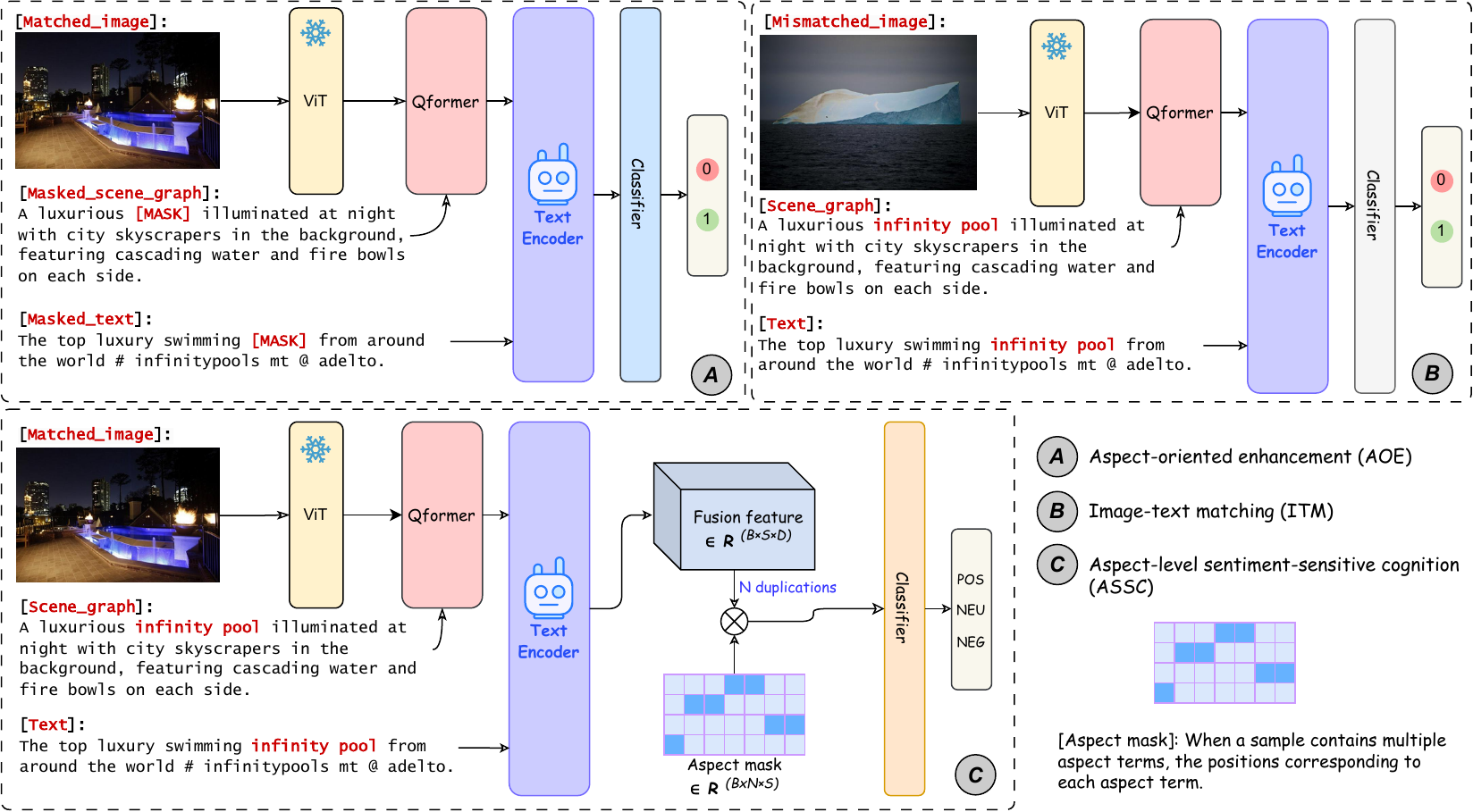}
    \caption{Continue pretraining for base model. We introduce three pre-training tasks—AOE, ITM, and ASSC—to enhance the model's aspect sensitivity, image-text alignment capability, and sentiment perception ability, respectively. In AOE and ITM tasks, 0 and 1 indicate whether the aspect entities in the image-text pair are aligned and whether the image-text features match. POS, NEU, and NEG represent Positive, Neutral, and Negative.}
    \label{fig:dasco3}
\end{figure*}

\begin{itemize}
    \item \textbf{AOE pairs} $(P_a,N_a)$: For each sample, we mask the aspect term $A$ in both $T$ and $S$ using a special token \verb|[Mask]|, yielding $(T_N, S_N)$. The original triplet $(T, V, S)$ forms a positive sample $P_a$ (label $L=1$), while the masked version $(T_N, V, S_N)$ is treated as a negative sample $N_a$ (label $L=0$).
    \item \textbf{ITM pairs} $(P_b,N_b)$: A negative sample $N_b = (T, V_N, S)$ (label $L=0$) is constructed by replacing the original image $V$ with a randomly sampled mismatched image $V_N$. The original triplet $(T, V, S)$ serves as the positive sample $P_b$ (label $L=1$).
    \item \textbf{ASSC quintuplets} $(T,V,S,A,y_c)$: Each sample has a full quintuplet used for aspect‑level sentiment classification with an aspect mask tensor.
\end{itemize}

Further implementation details are provided in the \textit{Supplementary Materials}.

\subsection{Continue Pretraining for Base Model}
\subsubsection{Aspect-Oriented Enhancement} Building upon the aspect-sensitive design proposed in \cite{DBLP:conf/ijcai/ZhuSGY024}, we propose an improved AOE task (Figure~\ref{fig:dasco3}.A) that explicitly trains the model to recognize the consistency of aspect information in input pairs. The model is trained to distinguish $P_a$ (with aspect) and $N_a$ (masked) with the following binary cross-entropy (BCE) loss:
\begin{equation}
    \centering
    \mathcal{L}_{AOE}=-\left[L\log(\hat{L})+(1-L)\log(1-\hat{L})\right],
    \label{eqLossaoe}
\end{equation}
where $\hat{L}$ denotes the predicted result.

\subsubsection{Image-Text Matching} To improve multimodal alignment, we adopt a standard ITM task (Figure~\ref{fig:dasco3}.B), following previous work \cite{DBLP:conf/acl/LingYX22, DBLP:conf/aaai/PengLWZZ24}.
Given a positive pair $P_b=(T, V, S)$ and a negative pair $N_b=(T, V_N, S)$, the model learns to predict alignment labels using the following BCE loss:
\begin{equation}
    \centering
    \mathcal{L}_{ITM}=-\left[L\log(\hat{L})+(1-L)\log(1-\hat{L})\right].
    \label{eqLossitm}
\end{equation}

\subsubsection{Aspect-level Sentiment-Sensitive Cognition} To address the SCP challenge by improving sentiment cue perception, we propose ASSC (Figure~\ref{fig:dasco3}.C), which performs aspect-level sentiment prediction.  
Given a sample with $N$ aspect terms, we replicate the fusion representation $H\in\mathbb{R}^{B\times S\times D}$ $N$ times to form a tensor of shape $\mathbb{R}^{B \times N \times S \times D}$.  
We then apply an aspect mask of shape $\mathbb{R}^{B\times N \times S}$, similar to an attention mask, to guide the model to focus on relevant features for each aspect. Finally, we predict the sentiment polarity probability $\hat{y}_i^{c}$ for each aspect. The training objective of ASSC is
\begin{equation}
    \centering
    \mathcal{L}_{ASSC}=-\frac{1}{N}\sum_{i=1}^{N}\log\hat{y}_{i}^c.
    \label{eqLossassc}
\end{equation}

\subsubsection{Joint Objective} The overall pretraining loss combines the original Qformer loss $\mathcal{L}_Q$ and the three subtask losses:
\begin{equation}
    \centering
    \mathcal{L}_{p} = \mathcal{L}_{Q} + \mathcal{L}_{AOE} + \mathcal{L}_{ITM} + \mathcal{L}_{ASSC}.
    \label{eqmulti}
\end{equation}

\subsection{Dependency Structure Augmented Contextual Scoping Module}
\begin{figure*}[t]
    \centering
    \includegraphics[width=1\linewidth]{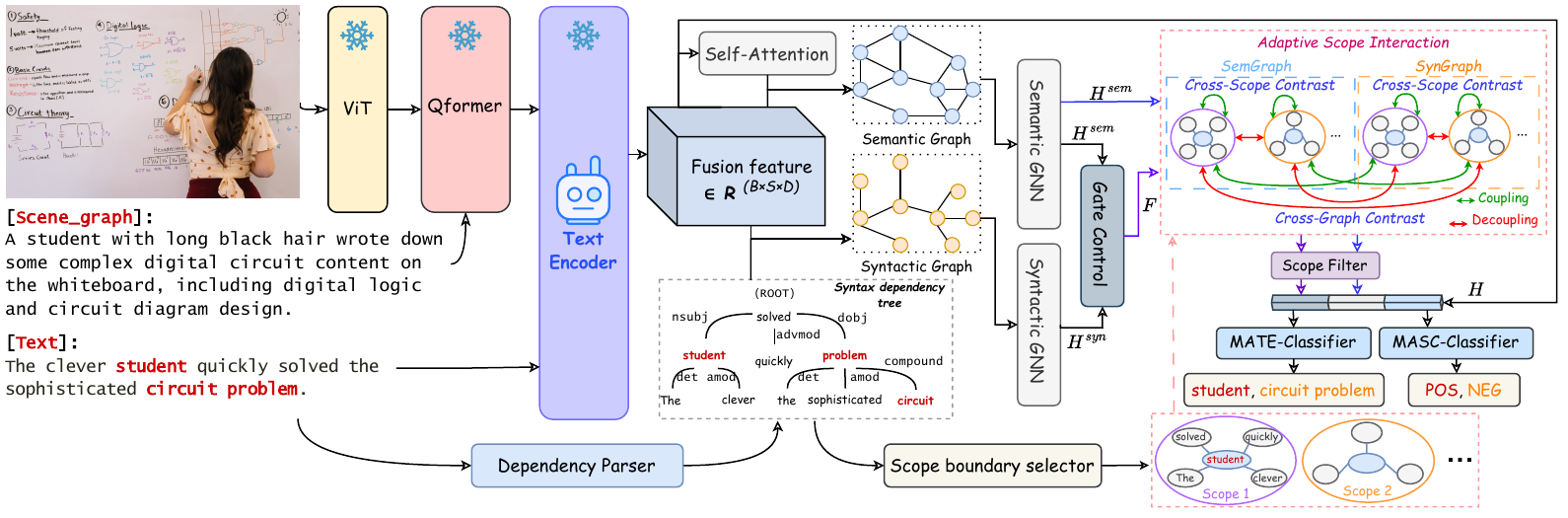}
    \caption{The pipeline of DASCO.}
    \label{fig:dasco5}
\end{figure*}
After base model continue pretraining, we freeze its parameters and introduce a dependency structure augmented contextual scoping module (Figure~\ref{fig:dasco5}), which integrates a syntactic-semantic dual-branch architecture with an adaptive scope
interaction mechanism to enhance context modeling. This module defines target-specific scopes based on the dependency tree, enabling the model to focus on key semantics within the defined context during reasoning.

\subsubsection{Target-specific Scope Definition}
\begin{figure}[t]
    \centering
    \includegraphics[width=1\linewidth]{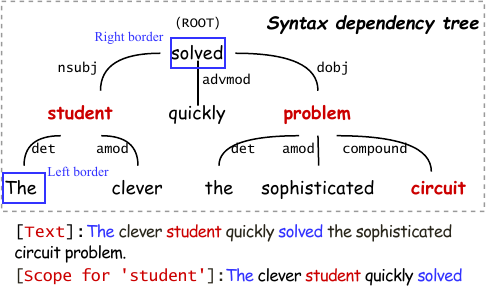}
    \caption{The definition of target-specific scope.}
    \label{fig:dasco4}
\end{figure}
To precisely delineate target-relevant words, we define a target-specific scope based on dependency structure derived from dependency parsing using spaCy\footnote{spaCy toolkit: \url{https://spacy.io/}}. For each target word, its scope spans from the leftmost to the rightmost word along its parent-child path in the dependency tree.
For instance, in ``The clever student quickly solved the sophisticated circuit problem,'' the scope of ``student'' spans from ``The'' to ``solved'', capturing not only direct dependents but also important modifiers such as ``quickly,'' as illustrated in Figure~\ref{fig:dasco4}. This approach effectively captures semantically relevant context.

\subsection{Graph Construction}
To incorporate both syntactic structure and semantic information from input sentences, we construct two graph-based representations: a semantic graph (SemGraph) and a syntactic graph (SynGraph).

\subsubsection{Semantic Graph Construction} Following \cite{DBLP:conf/emnlp/ChaiYTWNF023}, each word in sentence $S$ is treated as a node, and a directed graph is built using self-attention scores as edge weights. The resulting SemGraph captures semantic dependencies between word pairs. To model semantic interactions, we introduce a semantic graph neural network (SemGNN) for feature aggregation, producing enriched semantic representations $H^{sem}$:
\begin{equation}
    H_l^{sem}=\mathrm{ReLU}(A^{sem}H_{l-1}^{sem}W_l^{sem}+b_l^{sem}),
    \label{semG}
\end{equation}
where the adjacency matrix $A^{sem}\in\mathbb{R}^{S\times S}$ is computed via scaled dot-product attention:
\begin{equation}
    A^{sem}=\mathrm{softmax}\left(\frac{(QW_q)(KW_k)^T}{\sqrt{D}}\right),
    \label{semadj}
\end{equation}
with $Q, K \in \mathbb{R}^{S \times D}$ as duplicate contextual features from the text encoder. $W_q, W_k \in \mathbb{R}^{D \times D}$ are learnable projections. $D$ represents the dimension of each node feature, and $S$ is the sentence sequence length.
The initial input $H_0^{sem}$ is the encoder output $H$. Parameters for layer $l$ include the weight matrix $W_l^{sem} \in \mathbb{R}^{D \times D}$ and the bias term $b_l^{sem} \in \mathbb{R}^{D}$.

\subsubsection{Syntactic Graph Construction} We construct undirected SynGraph by parsing the sentence $S$ into a dependency tree $\mathbf{T}$ using spaCy. Each word serves as a node, and edges are formed based on dependency relationships. To encode global syntactic structure, we apply a syntactic graph neural network (SynGNN) to obtain syntactic representations $H^{syn}$:
\begin{equation}
    H_l^{syn}=\mathrm{ReLU}(A^{syn}H_{l-1}^{syn}W_l^{syn}+b_l^{syn}),
    \label{synG}
\end{equation}
where the adjacency matrix $A^{syn}\in\mathbb{R}^{S\times S}$ is defined as:
\begin{equation}
    A_{ij}^{syn} = 
    \begin{cases}
        1 & \text{if } \mathrm{adj}(i,j)=1 \text{ or } i=j, \\
        0 & \text{otherwise}.
    \end{cases}
    \label{adj}
\end{equation}

Here, $\mathrm{adj}(i, j)$ indicates the existence of a parent-child dependency relationship between word $i$ and word $j$. SynGNN uses the same formulation as SemGNN, but with independent parameters.

\subsubsection{Adaptive Scope Interaction via Contrastive Graph Learning}
We propose an adaptive scope interaction mechanism to reduce noise from irrelevant relationships and over-attended nodes in SynGraph and SemGraph. 
Our method precisely improves target-opinion alignment by distinguishing in-scope and out-of-scope words, effectively alleviating the SNE problem. 
It uses two contrastive strategies: (1) intragraph \textit{cross-scope} contrast, and (2) \textit{cross-graph} contrast.

To enhance alignment between target words and their corresponding opinion expressions, we introduce a \textit{cross-scope} contrastive loss based on node representations from SynGraph and SemGraph. Taking SynGraph as an example, let $\{n_i\}_{i\in[S]}$ represent all node representations, and $Scope_{t}^n$ the set of in-scope node for target $t$. The target node $n_t$ serves as the anchor: any node $n_{j}\in Scope_t^n$ forms a \textit{positive} pair with $n_t$, while nodes $n_{k}\notin Scope_t^n$ are treated as \textit{negatives}. Based on the InfoNCE \cite{DBLP:journals/corr/abs-1807-03748} objective, we compute the contrastive loss for all positive pairs $(n_t, n_{j})$ in the sentence as follows:
\begin{equation}
    \mathcal{L}^{syn}_{\mathrm{scope}}=-\frac{1}{T}\sum_{t=1}^T\sum_{n_j\in Scope_t^n}\log\frac{\exp(\mathrm{sim}(n_t,n_j)/\tau)}{\sum_{n_i}\exp(\mathrm{sim}(n_t,n_i)/\tau)},
    \label{syn intra}
\end{equation}
where $\tau$ denotes the temperature parameter, $T$ is the number of target words in the sentence, and $\mathrm{sim}(\cdot,\cdot)$ refers to cosine similarity. In parallel, we compute a corresponding contrastive loss $\mathcal{L}^{sem}_{\mathrm{scope}}$ for SemGraph. This objective promotes alignment with in-scope nodes while contrasting out-of-scope context, thereby reducing noise. The overall \textit{cross-scope} contrastive objective is $\mathcal{L}_{\mathrm{scope}}=\mathcal{L}^{syn}_{\mathrm{scope}}+\mathcal{L}^{sem}_{\mathrm{scope}}$.

To further enhance target-specific representations and align semantic scopes across heterogeneous graphs, we introduce a cross-graph contrastive learning strategy that integrates syntactic and semantic correlations. 
Taking SynGraph as an example, we designate the target node $n_t$ as the anchor. Its counterpart $m_t$ in SemGraph, along with all in-scope nodes $Scope_t^m$, constitute the \textit{positive} set, while all remaining SemGraph nodes are treated as \textit{negatives}. 
To emphasize differences among positives and boost representational diversity, we introduce a similarity weight $\omega = \mathrm{sim}(n_t, m_t)$ between anchor and its aligned node. The cross-graph contrastive loss for SynGraph is defined as follows:
\begin{equation}
    \mathcal{L}_{m_j}=-\log\frac{\exp(\omega/\tau)+\exp(\omega\cdot\mathrm{sim}(n_t,m_j)/\tau)}{\sum_{m_i}\exp(\mathrm{sim}(n_t,m_i)/\tau)},
    \label{graphlossmj}
\end{equation}
\begin{equation}
    \mathcal{L}^{syn}_{\mathrm{graph}} = \frac{1}{T}\sum_{t=1}^T\sum_{m_j\in Scope_t^m} \mathcal{L}_{m_j}.
    \label{graphloss}
\end{equation}

This objective aligns semantic and syntactic representations, and captures complementary information. 
We calculate the counterpart loss $\mathcal{L}^{sem}_{\mathrm{graph}}$ for SemGraph. The overall \textit{cross-graph} contrastive objective is $\mathcal{L}_{\mathrm{graph}}=\mathcal{L}^{syn}_{\mathrm{graph}}+\mathcal{L}^{sem}_{\mathrm{graph}}$.

Eventually, the overall loss for adaptive scope interaction is defined by combining cross-scope and cross-graph contrastive objectives:
\begin{equation}
\mathcal{L}_{\mathrm{ASI}}=\mathcal{L}_{\mathrm{scope}}+\mathcal{L}_{\mathrm{graph}}.
\label{cl}
\end{equation}

\subsubsection{Model Inference and Training}
During \textit{inference}, semantic and syntactic features ($H^{sem}$ and $H^{syn}$) are fused via a gated control mechanism to obtain $F$. A scope filter, implemented by a scope mask $\in \mathbb{R}^{B\times N\times S}$, functioning analogously to an attention mask, is applied to $F$ and $H^{sem}$ to selectively retain information within each of the $N$ target-specific scopes. The filtered representations are concatenated with the original encoding $H$ and fed into a task-specific classifier for the final prediction.

During \textit{training}, we employ task-specific strategies for each MABSA subtask. 
For the MATE task, \textit{nouns} and \textit{pronouns} are extracted via spaCy as candidate aspect terms (i.e., target words), and a scope is defined for each. A binary classifier then predicts whether each candidate is a genuine aspect term. The overall training objective is defined as:
\begin{equation}
\mathcal{L}_{\mathrm{MATE}} = - \frac{1}{N} \sum_{i=1}^{N} \mathrm{BCE}(y_i, \hat{y}_i) + \lambda \cdot \mathcal{L}_{\mathrm{ASI}},
\label{lossmate}
\end{equation}
where $\mathrm{BCE}(\cdot)$ denotes the binary cross-entropy loss, $y_i \in \{0,1\}$ is the ground-truth label, $\hat{y}_i$ is the predicted probability, $N$ is the number of candidates, and $\lambda$ is a hyperparameter.

For the MASC task, each identified aspect term (i.e., target word) is assigned a contextual scope, within which a three-way classifier predicts its sentiment polarity. The overall training objective is formulated as:
\begin{equation}
\mathcal{L}_{\mathrm{MASC}} = -\frac{1}{M}\sum_{j=1}^{M}\log\hat{y}_{j}^{y_j} + \lambda \cdot \mathcal{L}_{\mathrm{ASI}},
\label{lossmasc}
\end{equation}
where $y_j \in \{0,1,2\}$ is the ground truth sentiment label of the $j$-th aspect term, $\hat{y}_j^{y_j}$ is the predicted probability for the GT class $y_j$, and $M$ is the total number of aspect terms.

For the JMASA task, the trained MATE and MASC models are sequentially combined for inference without further training.

\section{Experiments}
\subsection{Settings}
\subsubsection{Datasets and Metrics}
Following previous work, we use Twitter2015 and Twitter2017 as benchmark datasets and build a new Scene Graph dataset based on them. Continue pretraining data are derived from these Twitter corpora (details in \textit{Supplementary Materials}). We evaluate performance using precision (P), recall (R), and Micro-F1 (F1) for MATE and JMASA tasks, and accuracy (Acc) and Macro-F1 (F1) for MASC.

\subsubsection{Implementation Details}
DASCO employs Qformer \cite{DBLP:conf/icml/0008LSH23} and FUSIE-base \cite{DBLP:conf/acl/PengLZ0Z23} as the base models, both initialized from the pretrained DQPSA \cite{DBLP:conf/aaai/PengLWZZ24}. The visual encoder is ViT \cite{DBLP:conf/icml/RadfordKHRGASAM21, DBLP:conf/iclr/DosovitskiyB0WZ21} (CLIP-ViT-bigG-14-laion2B-39B-b160k). 
Pre-training utilizes a learning rate of 5e-5, batch size of 4, over 50 epochs. In the fine-tuning phase, base model parameters are frozen while training proceeds with a learning rate of 2e-5, batch size of 16 for 20 epochs, with hyperparameter $\lambda = 0.2$. Both SemGNN and SynGNN employ $l=4$ layers. Training is conducted via distributed computing across 8 NVIDIA GeForce RTX 4090 GPUs. The code and checkpoints will be released upon publication.

\subsection{Main results}
We report the performance of DASCO and baselines on the JMASA, MATE, and MASC tasks in Tables \ref{tab:jmasa},\ref{tab:mate},\ref{tab:masc}. \textbf{Bolded} and \underline{underlined} values indicate the best and second best performance, respectively. 

\begin{table}[t]
\centering 
\setlength{\tabcolsep}{1mm}
\begin{tabular}{@{}lccccccc@{}}
\toprule 
\multicolumn{1}{l}{\multirow{2}{*}{Method}}  & \multicolumn{3}{c}{Twitter2015} & \multicolumn{3}{c}{Twitter2017} \\
\cmidrule(lr){2-4} \cmidrule(lr){5-7}
\multicolumn{1}{c}{} & P & R & F1 & P & R & F1 \\
\midrule 
JML\shortcite{DBLP:conf/emnlp/JuZXLLZZ21} & 65.0 & 63.2 & 64.1 & 66.5 & 65.5 & 66.0 \\
VLP\shortcite{DBLP:conf/acl/LingYX22} & 65.1 & 68.3 & 66.6 & 66.9 & 69.2 & 68.2 \\
CMMT\shortcite{DBLP:journals/ipm/YangNY22} & 64.6 & 68.7 & 66.5 & 67.6 & 69.4 & 68.5 \\
M2DF\shortcite{DBLP:conf/emnlp/ZhaoLWOZD23} & 67.0 & 67.3 & 67.6 & 67.9 & 68.8 & 68.3 \\
AoM\shortcite{DBLP:conf/acl/ZhouGLYZY23} & 67.9 & 69.3 & 68.6 & 68.4 & 71.0 & 69.7 \\
MOCOLNET\shortcite{DBLP:journals/tkde/MuNWXZL24} & 66.3 & 67.9 & 67.1 & 67.3 & 68.7 & 68.0 \\
DQPSA\shortcite{DBLP:conf/aaai/PengLWZZ24} & \underline{71.7} & 72.0 & 71.9 & 71.1 & 70.2 & 70.6 \\
Atlantis\shortcite{DBLP:journals/inffus/XiaoWXLJH24} & 65.6 & 69.2 & 67.3 & 68.6 & 70.3 & 69.4 \\
ADAR\shortcite{DBLP:conf/mm/ChenZXZ0024} & 70.0 & 71.5 & 71.2 & \textbf{71.6} & 71.0 & 71.4 \\
AESAL\shortcite{DBLP:conf/ijcai/ZhuSGY024} & 68.7 & 70.4 & 69.5 & 69.4  & \textbf{74.8} & \textbf{72.0} \\
RNG\shortcite{DBLP:conf/icmcs/LiuZLZSZH24} & 67.8 & 69.5 & 68.6 & 69.5 & 71.0 & 70.2 \\
Vanessa\shortcite{DBLP:conf/emnlp/Xiao0ZHC24} & 68.6 & 71.1 & 69.8 & 69.2 & 72.1 & 70.6 \\
CORSA\shortcite{DBLP:conf/coling/LiuLYZW25} & 69.0 & 70.8 & 69.9 & 70.1 & 71.0 & 70.6 \\
DEQA\shortcite{DBLP:conf/aaai/HanHBWL25} & 71.4 & \underline{73.9} & \underline{72.7} & \underline{71.4} & 72.4 & \underline{71.9} \\
\textbf{DASCO}  & \textbf{72.7} & \textbf{77.4} & \textbf{75.0} & 70.6 & \underline{73.3} & \textbf{72.0} \\
\bottomrule 
\end{tabular}
\caption{Performance comparison on JMASA task.}
\label{tab:jmasa}
\end{table}

\subsubsection{Results for JMASA} 
On Twitter2015, DASCO surpasses the second-best model DQPSA by 3.1 F1 points, primarily attribute to its MABSA-specific pretraining strategy and efficient denoising mechanism that jointly enhance aspect and sentiment reasoning.
On Twitter2017, although ADAR and AESAL achieve marginally higher precision and recall, DASCO still yields the highest overall F1 score. 
Its superior performance is attributed to three core components: (i) a multitask continue pretraining scheme designed to enhance the model's aspect sensitivity, image-text alignment, and sentiment perception; (ii) an efficient integration of syntactic and semantic information; and (iii) a target-specific scoping mechanism with adaptive scope interaction, which effectively filters semantic noise from irrelevant regions.

\subsubsection{Results for MATE and MASC} Tables~\ref{tab:mate} and~\ref{tab:masc} report DASCO's results on MATE and MASC tasks. Consistent with JMASA, DASCO achieves state-of-the-art performance on most metrics across both tasks. The sole exception is on Twitter2015-MATE, where DASCO shows marginally lower precision than DQPSA and DEQA, while still achieves the highest F1 score. These findings further validate our approach's robustness.

\begin{table}[t]
\centering
\setlength{\tabcolsep}{1mm}
\begin{tabular}{@{}lccccccc@{}}
\toprule
\multicolumn{1}{l}{\multirow{2}{*}{Method}} & \multicolumn{3}{c}{Twitter2015} & \multicolumn{3}{c}{Twitter2017} \\
\cmidrule(lr){2-4} \cmidrule(lr){5-7}
\multicolumn{1}{c}{} & P & R & F1 & P & R & F1 \\
\midrule
JML\shortcite{DBLP:conf/emnlp/JuZXLLZZ21} & 83.6 & 81.2 & 82.4 & 92.0 & 90.7 & 91.4 \\
VLP\shortcite{DBLP:conf/acl/LingYX22} & 83.6 & 87.9 & 85.7 & 90.8 & 92.6 & 91.7 \\
CMMT\shortcite{DBLP:journals/ipm/YangNY22} & 83.9 & 88.1 & 85.9 & 92.2 & 93.9 & 93.1 \\
MNER-QG\shortcite{DBLP:conf/aaai/JiaSSL00CL23} & 77.4 & 72.1 & 74.7 & 88.2 & 85.6 & 86.9 \\
Prompt-Me-Up\shortcite{DBLP:conf/mm/HuCLMWY23} & 80.0 & 80.9 & 80.5 & 91.7 & 91.3 & 91.6 \\
M2DF\shortcite{DBLP:conf/emnlp/ZhaoLWOZD23} & 85.0 & 87.2 & 86.1 & 91.2 & 93.0 & 92.2 \\
AoM\shortcite{DBLP:conf/acl/ZhouGLYZY23} & 84.6 & 87.9 & 86.2 & 91.8 & 92.8 & 92.3 \\
MOCOLNET\shortcite{DBLP:journals/tkde/MuNWXZL24} & 85.3 & 87.0 & 86.1 & 91.5 & 92.9 & 91.7 \\
DQPSA\shortcite{DBLP:conf/aaai/PengLWZZ24} &  \textbf{88.3} & 87.1 & 87.7 & \underline{95.1} & 93.5 & 94.3 \\
Atlantis\shortcite{DBLP:journals/inffus/XiaoWXLJH24} & 84.2 & 87.7 & 86.1 & 91.8 & 93.2 & 92.7 \\
ADAR\shortcite{DBLP:conf/mm/ChenZXZ0024} & 86.5 & 88.0 & 87.2 & 93.0 & 93.9 & 93.4 \\
CORSA\shortcite{DBLP:conf/coling/LiuLYZW25} & 85.1 & 87.6 & 86.3 & 92.6 & 93.0 & 92.8 \\
DEQA\shortcite{DBLP:conf/aaai/HanHBWL25} & \underline{86.6} & \underline{89.5} & \underline{88.0} & 93.8 & \underline{95.1} & \underline{94.4} \\
\textbf{DASCO} & 86.5 & \textbf{92.3} & \textbf{89.2} & \textbf{95.6} & \textbf{96.1} & \textbf{95.8} \\
\bottomrule
\end{tabular}
\caption{Performance comparison on MATE task.}
\label{tab:mate}
\end{table}

\begin{table}[t]
\centering
\begin{tabular}{@{}llccccc@{}}
\toprule
\multicolumn{1}{l}{\multirow{2}{*}{Method}} & \multicolumn{2}{c}{Twitter2015} & \multicolumn{2}{c}{Twitter2017} \\
\cmidrule(lr){2-3} \cmidrule(lr){4-5}
\multicolumn{1}{c}{} &ACC  &F1 &ACC  &F1 \\
\midrule
JML\shortcite{DBLP:conf/emnlp/JuZXLLZZ21} & 76.1 & 72.3 & 72.3 & 69.9  \\
VLP\shortcite{DBLP:conf/acl/LingYX22} & 78.6 & 73.8 & 73.8 & 71.8  \\
CMMT\shortcite{DBLP:journals/ipm/YangNY22} & 77.9 & - & 73.8 & -  \\
M2DF\shortcite{DBLP:conf/emnlp/ZhaoLWOZD23} & 78.9 & 74.8 & 74.3 & 73.0  \\
AoM\shortcite{DBLP:conf/acl/ZhouGLYZY23} & 80.2 & 75.9 & 76.4 & 75.0  \\
DQPSA\shortcite{DBLP:conf/aaai/PengLWZZ24} &  81.1 & - & 75.0 & -  \\
Atlantis\shortcite{DBLP:journals/inffus/XiaoWXLJH24} & 79.3 & - & 74.2 & -  \\
ADAR\shortcite{DBLP:conf/mm/ChenZXZ0024}  & 81.3 & 77.1 & 77.2 & \underline{76.6}  \\
AESAL\shortcite{DBLP:conf/ijcai/ZhuSGY024} & 80.1 & 75.2 & \underline{78.8} & 75.9 \\
CORSA\shortcite{DBLP:conf/coling/LiuLYZW25} & 81.1 & \underline{77.7} & 76.6 & 74.5  \\
DEQA\shortcite{DBLP:conf/aaai/HanHBWL25} & \underline{82.1} & 77.6 & 75.8 & 75.1 \\
\textbf{DASCO} &\textbf{83.8} & \textbf{78.8} & \textbf{79.5} & \textbf{76.9}  \\
\bottomrule
\end{tabular}%
\caption{Performance comparison on MASC task.}
\label{tab:masc}
\end{table}

\subsection{Ablation Study}
We conduct an ablation study across five key components of DASCO to evaluate their individual contributions, as summarized in Table~\ref{tab:ab}:

\begin{table*}[t]
\centering
\setlength{\tabcolsep}{1.5mm}
\begin{tabular}{lcccccccccccccccc}
\toprule
\multirow{3}{*}{Variants} & \multicolumn{6}{c}{JMASA} & \multicolumn{6}{c}{MATE} & \multicolumn{4}{c}{MASC} \\ \cmidrule(lr){2-7} \cmidrule(lr){8-13} \cmidrule(lr){14-17}
& \multicolumn{3}{c}{Twitter2015} & \multicolumn{3}{c}{Twitter2017} & \multicolumn{3}{c}{Twitter2015} & \multicolumn{3}{c}{Twitter2017} & \multicolumn{2}{c}{Twitter2015} & \multicolumn{2}{c}{Twitter2017} \\ \cmidrule(lr){2-7} \cmidrule(lr){8-13} \cmidrule(lr){14-17}
 & P & R & F1 & P & R & F1 & P & R  & F1  & P & R & F1  & ACC & F1  & ACC & F1             \\
\midrule
Full & \textbf{72.7} & \textbf{77.4}  & \textbf{75.0} & \textbf{70.6} & \textbf{73.3} & \textbf{72.0} & \textbf{86.4} & 92.3 & \textbf{89.2}  & \textbf{95.6}  & 96.1 & \textbf{95.8} & 83.8  & \textbf{78.8} & \textbf{79.5}  & \textbf{76.9}  \\
w/o Scene graph  & 68.1 & 77.1 & 72.3 & 67.9  & 72.4  & 70.1 & 81.5  & \textbf{92.4}  & 86.6  & 91.1  & 97.1  & 94.0 & \textbf{83.9}   & 78.3   & 78.8 & 76.3  \\
w/o Scope    & 66.9 & 75.5 & 70.9 & 66.1 & 72.3 & 69.1 & 81.4 & 92.1 & 86.4 & 89.0 & 97.2 & 93.0 & 82.7 & 77.2 & 72.1 & 70.8 \\
w/o Syntactic & 54.9  & 61.8 & 58.1 & 61.7 & 68.7 & 65.0 & 67.9 & 71.5 & 69.7 & 76.0 & 86.3 & 80.8 & 75.1 & 71.2 & 68.3 & 65.2 \\
w/o Semantic & 69.1 & 76.3 & 72.5  & 61.8 & 67.0 & 64.3 & 86.2 & 91.2 & 86.7 & 90.1 & \textbf{97.6} & 93.7 & 82.2 & 77.4 & 75.1 & 73.3   \\
w/o Pretraining & 58.4 & 63.7 & 61.0 & 60.8 & 64.3 & 62.5 & 84.6 & 91.6 & 88.0 & 92.0 & 96.7 & 94.3 & 75.5 & 72.1 & 67.3   & 64.6 \\
\bottomrule
\end{tabular}
\caption{The performance comparison of our full model and its ablated methods on JMASA, MATE and MASC.}
\label{tab:ab}
\end{table*}

\subsubsection{Effect of Scene graph}\textit{W/o Scene graph} replaces structured scene graphs with raw text input to Qformer. This leads to notable performance drops in most evaluation metrics across all tasks, except for Recall in MATE and Accuracy in MASC. These findings indicate that structured image descriptions improve Qformer's textual-visual alignment and enhance aspect-level understanding.

\subsubsection{Effect of Scope}\textit{W/o Scope} substitutes the original scope with only the target word, resulting in substantial performance. This validates our scope construction strategy's effectiveness in modeling semantic boundaries.

\subsubsection{Effect of Syntactic branch}\textit{W/o Syntactic} removes the syntactic branch, retaining only the SemGraph representation $H^{sem}$ and its cross-scope contrast within the adaptive scope interaction. The notable performance decline highlights the pivotal role of syntactic information in scope-level denoising and further affirms the effectiveness of cross-graph contrast in enhancing semantic interaction quality.

\subsubsection{Effect of Semantic branch}\textit{W/o Semantic} liminates the semantic branch, similar to how \textit{W/o Syntactic} removes the syntactic branch. The moderate performance drop demonstrates the syntactic branch's effectiveness in capturing sentiment cues and its key role in denoising, while indicating SemGraph's quality limitations.

\subsubsection{Effect of Continue pretraining}\textit{W/o Pretraining} omits continue pretraining and directly fine-tunes the full model. The results demonstrate performance degradation across all tasks, with particularly pronounced drops in the MASC task. This highlights the effectiveness of our continue pretraining strategy in enhancing aspect and sentiment reasoning, thereby mitigating the SCP challenge.

\subsection{Comparison with LLMs on MASC}
To validate DASCO's competitiveness in MABSA, we compared it with mainstream LLMs including VisualGLM-6B, GPT-3.5, and GPT-4V. Given LLMs' inherent limitations in aspect identification and structured output, comparisons were restricted to the MASC task for evaluation fairness.
We defined roles and system instructions for each LLM and provided image, text, and aspect information as input (GPT-3.5 received only text and aspect inputs due to its unimodal nature). Results in Table \ref{tab:llm} demonstrate that despite DASCO's parameter significantly smaller parameter scale compared to LLMs, it exhibits superior performance, confirming its effectiveness in multimodal sentiment understanding tasks.

\begin{table}[t]
\centering
\begin{tabular}{lcccc}
\toprule
\multirow{2}{*}{Models} & \multicolumn{2}{c}{Twitter2015} & \multicolumn{2}{c}{Twitter2017} \\ \cmidrule(lr){2-3} \cmidrule(lr){4-5}
& ACC & F1 & ACC & F1             \\
\midrule
VisualGLM-6B  & 66.7 & 67.2 & 68.1 & 67.9 \\
GPT3.5 & 65.3 & 66.6 & 67.0 & 67.1  \\
GPT4V  & 75.6 & 74.3 & 75.5 & 75.2   \\
\textbf{DASCO (ours)}  & \textbf{83.8} & \textbf{78.8} & \textbf{79.5} & \textbf{76.9}   \\
\bottomrule
\end{tabular}
\caption{Comparison with LLMs on MASC task.}
\label{tab:llm}
\end{table}

\subsection{Computation Efficiency}
Furthermore, we analyzed the computational efficiency of DASCO compared to several classical models under consistent settings. As shown in Table \ref{tab:comput}, DASCO demonstrates superior inference speed on the Twitter2015/2017 test sets. During inference, DASCO avoids complex attention mechanisms \cite{DBLP:conf/emnlp/JuZXLLZZ21, DBLP:conf/acl/ZhouGLYZY23}, eliminates calculations of numerous atomic feature similarities \cite{DBLP:conf/acl/ZhouGLYZY23}, and differs from autoregressive decoding architectures like BART \cite{DBLP:conf/acl/LingYX22}. Instead, it employs only two concise GNN modules, significantly enhancing inference efficiency while maintaining performance.

\begin{table}[t]
\centering
\begin{tabular}{lcc}
\toprule
Methods              & Twitter2015     & Twitter2017     \\ \midrule
JML\shortcite{DBLP:conf/emnlp/JuZXLLZZ21}                  & 22.390s           & 17.507s          \\
VLP\shortcite{DBLP:conf/acl/LingYX22}                  & 29.297s          & 18.165s          \\
AoM\shortcite{DBLP:conf/acl/ZhouGLYZY23}                  & 30.609s          & 46.481s          \\
DQPSA\shortcite{DBLP:conf/aaai/PengLWZZ24}                & 80.943s          & 93.177s          \\
\textbf{DASCO(ours)} & \textbf{14.248s} & \textbf{11.108s} \\
\bottomrule
\end{tabular}
\caption{Comparison on computation efficiency.}
\label{tab:comput}
\end{table}

\section{Conclusion}
This paper proposes a Dependency Structure Augmented Contextual Scoping Framework (DASCO) that effectively addresses \textit{Multimodal Information Misalignment} (MIM) and \textit{Sentiment Cue Perception} (SCP) challenges through continue base model pretraining. The introduced syntactic-semantic dual-branch architecture, combined with the target-specific scope mechanism, leverages adaptive scope interaction to guide the model in suppressing irrelevant noise, thereby addressing the \textit{Semantic Noise Elimination} (SNE) challenge. 
Extensive experiments on two benchmark datasets demonstrate the superiority of DASCO, with significant improvements in JMASA task (+3.1\% F1 and +5.4\% precision on Twitter2015) and 37\% faster inference than the current fastest approach on both two datasets. Furthermore, this paper identifies three major challenges in MABSA, namely SCP, MIM, and SNE, which merit further investigation in future research.
\appendix

\bibliography{aaai2026}

\begin{thebibliography}{36}
\providecommand{\natexlab}[1]{#1}

\bibitem[{Chai et~al.(2023)Chai, Yao, Tang, Wang, Nie, Fang, and Liao}]{DBLP:conf/emnlp/ChaiYTWNF023}
Chai, H.; Yao, Z.; Tang, S.; Wang, Y.; Nie, L.; Fang, B.; and Liao, Q. 2023.
\newblock Aspect-to-Scope Oriented Multi-view Contrastive Learning for Aspect-based Sentiment Analysis.
\newblock In \emph{Findings of the Association for Computational Linguistics: {EMNLP} 2023, Singapore, December 6-10, 2023}, 10902--10913. Association for Computational Linguistics.

\bibitem[{Chen, Tian, and Song(2020)}]{DBLP:conf/coling/ChenTS20}
Chen, G.; Tian, Y.; and Song, Y. 2020.
\newblock Joint Aspect Extraction and Sentiment Analysis with Directional Graph Convolutional Networks.
\newblock In \emph{Proceedings of the 28th International Conference on Computational Linguistics, {COLING} 2020, Barcelona, Spain (Online), December 8-13, 2020}, 272--279. International Committee on Computational Linguistics.

\bibitem[{Chen et~al.(2024)Chen, Zhu, Xu, Zhang, Wu, and Zheng}]{DBLP:conf/mm/ChenZXZ0024}
Chen, Z.; Zhu, Z.; Xu, W.; Zhang, Y.; Wu, X.; and Zheng, Y. 2024.
\newblock \emph{Aspects are Anchors: } Towards Multimodal Aspect-based Sentiment Analysis via Aspect-driven Alignment and Refinement.
\newblock In \emph{Proceedings of the 32nd {ACM} International Conference on Multimedia, {MM} 2024, Melbourne, VIC, Australia, 28 October 2024 - 1 November 2024}, 2292--2300. {ACM}.

\bibitem[{Dai et~al.(2021)Dai, Yan, Sun, Liu, and Qiu}]{DBLP:conf/naacl/DaiYSLQ21}
Dai, J.; Yan, H.; Sun, T.; Liu, P.; and Qiu, X. 2021.
\newblock Does syntax matter? {A} strong baseline for Aspect-based Sentiment Analysis with RoBERTa.
\newblock In \emph{Proceedings of the 2021 Conference of the North American Chapter of the Association for Computational Linguistics: Human Language Technologies, {NAACL-HLT} 2021, Online, June 6-11, 2021}, 1816--1829. Association for Computational Linguistics.

\bibitem[{Devlin et~al.(2019)Devlin, Chang, Lee, and Toutanova}]{DBLP:conf/naacl/DevlinCLT19}
Devlin, J.; Chang, M.; Lee, K.; and Toutanova, K. 2019.
\newblock {BERT:} Pre-training of Deep Bidirectional Transformers for Language Understanding.
\newblock In \emph{Proceedings of the 2019 Conference of the North American Chapter of the Association for Computational Linguistics: Human Language Technologies, {NAACL-HLT} 2019, Minneapolis, MN, USA, June 2-7, 2019, Volume 1 (Long and Short Papers)}, 4171--4186. Association for Computational Linguistics.

\bibitem[{Dosovitskiy et~al.(2021)Dosovitskiy, Beyer, Kolesnikov, Weissenborn, Zhai, Unterthiner, Dehghani, Minderer, Heigold, Gelly, Uszkoreit, and Houlsby}]{DBLP:conf/iclr/DosovitskiyB0WZ21}
Dosovitskiy, A.; Beyer, L.; Kolesnikov, A.; Weissenborn, D.; Zhai, X.; Unterthiner, T.; Dehghani, M.; Minderer, M.; Heigold, G.; Gelly, S.; Uszkoreit, J.; and Houlsby, N. 2021.
\newblock An Image is Worth 16x16 Words: Transformers for Image Recognition at Scale.
\newblock In \emph{9th International Conference on Learning Representations, {ICLR} 2021, Virtual Event, Austria, May 3-7, 2021}. OpenReview.net.

\bibitem[{Feng et~al.(2024)Feng, Lin, Shang, and Gao}]{DBLP:conf/coling/FengL0G24}
Feng, J.; Lin, M.; Shang, L.; and Gao, X. 2024.
\newblock Autonomous Aspect-Image Instruction a2II: Q-Former Guided Multimodal Sentiment Classification.
\newblock In \emph{Proceedings of the 2024 Joint International Conference on Computational Linguistics, Language Resources and Evaluation, {LREC/COLING} 2024, 20-25 May, 2024, Torino, Italy}, 1996--2005. {ELRA} and {ICCL}.

\bibitem[{Han et~al.(2025)Han, Hu, Bai, Wang, and Luo}]{DBLP:conf/aaai/HanHBWL25}
Han, Z.; Hu, M.; Bai, Y.; Wang, X.; and Luo, B. 2025.
\newblock {DEQA:} Descriptions Enhanced Question-Answering Framework for Multimodal Aspect-Based Sentiment Analysis.
\newblock In \emph{AAAI-25, Sponsored by the Association for the Advancement of Artificial Intelligence, February 25 - March 4, 2025, Philadelphia, PA, {USA}}, 23987--23995. {AAAI} Press.

\bibitem[{Hu et~al.(2023)Hu, Chen, Liu, Meng, Wen, and Yu}]{DBLP:conf/mm/HuCLMWY23}
Hu, X.; Chen, J.; Liu, A.; Meng, S.; Wen, L.; and Yu, P.~S. 2023.
\newblock Prompt Me Up: Unleashing the Power of Alignments for Multimodal Entity and Relation Extraction.
\newblock In \emph{Proceedings of the 31st {ACM} International Conference on Multimedia, {MM} 2023, Ottawa, ON, Canada, 29 October 2023- 3 November 2023}, 5185--5194. {ACM}.

\bibitem[{Huang et~al.(2024)Huang, Xu, Li, Ye, and Lin}]{DBLP:conf/coling/HuangXLYL24}
Huang, S.; Xu, B.; Li, C.; Ye, J.; and Lin, X. 2024.
\newblock {MNER-MI:} {A} Multi-image Dataset for Multimodal Named Entity Recognition in Social Media.
\newblock In \emph{Proceedings of the 2024 Joint International Conference on Computational Linguistics, Language Resources and Evaluation, {LREC/COLING} 2024, 20-25 May, 2024, Torino, Italy}, 11452--11462. {ELRA} and {ICCL}.

\bibitem[{Jia et~al.(2023)Jia, Shen, Shen, Liao, Chen, He, Chen, and Li}]{DBLP:conf/aaai/JiaSSL00CL23}
Jia, M.; Shen, L.; Shen, X.; Liao, L.; Chen, M.; He, X.; Chen, Z.; and Li, J. 2023.
\newblock {MNER-QG:} An End-to-End {MRC} Framework for Multimodal Named Entity Recognition with Query Grounding.
\newblock In \emph{Thirty-Seventh {AAAI} Conference on Artificial Intelligence, {AAAI} 2023, Washington, DC, USA, February 7-14, 2023}, 8032--8040. {AAAI} Press.

\bibitem[{Ju et~al.(2021)Ju, Zhang, Xiao, Li, Li, Zhang, and Zhou}]{DBLP:conf/emnlp/JuZXLLZZ21}
Ju, X.; Zhang, D.; Xiao, R.; Li, J.; Li, S.; Zhang, M.; and Zhou, G. 2021.
\newblock Joint Multi-modal Aspect-Sentiment Analysis with Auxiliary Cross modal Relation Detection.
\newblock In \emph{Proceedings of the 2021 Conference on Empirical Methods in Natural Language Processing, {EMNLP} 2021, Virtual Event / Punta Cana, Dominican Republic, 7-11 November, 2021}, 4395--4405. Association for Computational Linguistics.

\bibitem[{Li et~al.(2023)Li, Li, Savarese, and Hoi}]{DBLP:conf/icml/0008LSH23}
Li, J.; Li, D.; Savarese, S.; and Hoi, S. C.~H. 2023.
\newblock {BLIP-2:} Bootstrapping Language-Image Pre-training with Frozen Image Encoders and Large Language Models.
\newblock In \emph{International Conference on Machine Learning, {ICML} 2023, 23-29 July 2023, Honolulu, Hawaii, {USA}}, volume 202 of \emph{Proceedings of Machine Learning Research}, 19730--19742. {PMLR}.

\bibitem[{Li et~al.(2024{\natexlab{a}})Li, Li, Sun, Wang, Zhang, Wang, and Pan}]{DBLP:conf/acl/LiLSWZW024}
Li, J.; Li, H.; Sun, D.; Wang, J.; Zhang, W.; Wang, Z.; and Pan, G. 2024{\natexlab{a}}.
\newblock LLMs as Bridges: Reformulating Grounded Multimodal Named Entity Recognition.
\newblock In \emph{Findings of the Association for Computational Linguistics, {ACL} 2024, Bangkok, Thailand and virtual meeting, August 11-16, 2024}, 1302--1318. Association for Computational Linguistics.

\bibitem[{Li et~al.(2024{\natexlab{b}})Li, Yu, Yang, Wang, Yang, and Xia}]{DBLP:conf/mm/LiYYWYX24}
Li, Z.; Yu, J.; Yang, J.; Wang, W.; Yang, L.; and Xia, R. 2024{\natexlab{b}}.
\newblock Generative Multimodal Data Augmentation for Low-Resource Multimodal Named Entity Recognition.
\newblock In \emph{Proceedings of the 32nd {ACM} International Conference on Multimedia, {MM} 2024, Melbourne, VIC, Australia, 28 October 2024 - 1 November 2024}, 7336--7345. {ACM}.

\bibitem[{Ling, Yu, and Xia(2022)}]{DBLP:conf/acl/LingYX22}
Ling, Y.; Yu, J.; and Xia, R. 2022.
\newblock Vision-Language Pre-Training for Multimodal Aspect-Based Sentiment Analysis.
\newblock In \emph{Proceedings of the 60th Annual Meeting of the Association for Computational Linguistics (Volume 1: Long Papers), {ACL} 2022, Dublin, Ireland, May 22-27, 2022}, 2149--2159. Association for Computational Linguistics.

\bibitem[{Liu, He, and Liang(2024)}]{DBLP:conf/icmcs/LiuHL24}
Liu, H.; He, L.; and Liang, J. 2024.
\newblock Joint Modal Circular Complementary Attention for Multimodal Aspect-Based Sentiment Analysis.
\newblock In \emph{{IEEE} International Conference on Multimedia and Expo, {ICME} 2024 - Workshops, Niagara Falls, ON, Canada, July 15-19, 2024}, 1--6. {IEEE}.

\bibitem[{Liu et~al.(2025)Liu, Li, Ye, Zhang, and Wang}]{DBLP:conf/coling/LiuLYZW25}
Liu, X.; Li, R.; Ye, S.; Zhang, G.; and Wang, X. 2025.
\newblock Multimodal Aspect-Based Sentiment Analysis under Conditional Relation.
\newblock In \emph{Proceedings of the 31st International Conference on Computational Linguistics, {COLING} 2025, Abu Dhabi, UAE, January 19-24, 2025}, 313--323. Association for Computational Linguistics.

\bibitem[{Liu et~al.(2019)Liu, Ott, Goyal, Du, Joshi, Chen, Levy, Lewis, Zettlemoyer, and Stoyanov}]{DBLP:journals/corr/abs-1907-11692}
Liu, Y.; Ott, M.; Goyal, N.; Du, J.; Joshi, M.; Chen, D.; Levy, O.; Lewis, M.; Zettlemoyer, L.; and Stoyanov, V. 2019.
\newblock RoBERTa: {A} Robustly Optimized {BERT} Pretraining Approach.
\newblock arXiv:1907.11692.

\bibitem[{Liu et~al.(2024)Liu, Zhou, Li, Zhang, Shang, Zhang, and Hu}]{DBLP:conf/icmcs/LiuZLZSZH24}
Liu, Y.; Zhou, Y.; Li, Z.; Zhang, J.; Shang, Y.; Zhang, C.; and Hu, S. 2024.
\newblock {RNG:} Reducing Multi-level Noise and Multi-grained Semantic Gap for Joint Multimodal Aspect-Sentiment Analysis.
\newblock In \emph{{IEEE} International Conference on Multimedia and Expo, {ICME} 2024, Niagara Falls, ON, Canada, July 15-19, 2024}, 1--6. {IEEE}.

\bibitem[{Ma et~al.(2023)Ma, Hu, Liu, Yang, Li, Yu, and Wen}]{DBLP:conf/acl/MaHLYLYW23}
Ma, F.; Hu, X.; Liu, A.; Yang, Y.; Li, S.; Yu, P.~S.; and Wen, L. 2023.
\newblock AMR-based Network for Aspect-based Sentiment Analysis.
\newblock In \emph{Proceedings of the 61st Annual Meeting of the Association for Computational Linguistics (Volume 1: Long Papers), {ACL} 2023, Toronto, Canada, July 9-14, 2023}, 322--337. Association for Computational Linguistics.

\bibitem[{Mu et~al.(2024)Mu, Nie, Wang, Xu, Zhang, and Liu}]{DBLP:journals/tkde/MuNWXZL24}
Mu, J.; Nie, F.; Wang, W.; Xu, J.; Zhang, J.; and Liu, H. 2024.
\newblock MOCOLNet: {A} Momentum Contrastive Learning Network for Multimodal Aspect-Level Sentiment Analysis.
\newblock \emph{{IEEE} Trans. Knowl. Data Eng.}, 36(12): 8787--8800.

\bibitem[{Peng et~al.(2024)Peng, Li, Wang, Zhang, and Zhao}]{DBLP:conf/aaai/PengLWZZ24}
Peng, T.; Li, Z.; Wang, P.; Zhang, L.; and Zhao, H. 2024.
\newblock A Novel Energy Based Model Mechanism for Multi-Modal Aspect-Based Sentiment Analysis.
\newblock In \emph{Thirty-Eighth {AAAI} Conference on Artificial Intelligence, {AAAI} 2024, February 20-27, 2024, Vancouver, Canada}, 18869--18878. {AAAI} Press.

\bibitem[{Peng et~al.(2023)Peng, Li, Zhang, Du, and Zhao}]{DBLP:conf/acl/PengLZ0Z23}
Peng, T.; Li, Z.; Zhang, L.; Du, B.; and Zhao, H. 2023.
\newblock {FSUIE:} {A} Novel Fuzzy Span Mechanism for Universal Information Extraction.
\newblock In \emph{Proceedings of the 61st Annual Meeting of the Association for Computational Linguistics (Volume 1: Long Papers), {ACL} 2023, Toronto, Canada, July 9-14, 2023}, 16318--16333. Association for Computational Linguistics.

\bibitem[{Radford et~al.(2021)Radford, Kim, Hallacy, Ramesh, Goh, Agarwal, Sastry, Askell, Mishkin, Clark, Krueger, and Sutskever}]{DBLP:conf/icml/RadfordKHRGASAM21}
Radford, A.; Kim, J.~W.; Hallacy, C.; Ramesh, A.; Goh, G.; Agarwal, S.; Sastry, G.; Askell, A.; Mishkin, P.; Clark, J.; Krueger, G.; and Sutskever, I. 2021.
\newblock Learning Transferable Visual Models From Natural Language Supervision.
\newblock In \emph{Proceedings of the 38th International Conference on Machine Learning, {ICML} 2021, 18-24 July 2021, Virtual Event}, volume 139 of \emph{Proceedings of Machine Learning Research}, 8748--8763. {PMLR}.

\bibitem[{Tang et~al.(2020)Tang, Ji, Li, and Zhou}]{DBLP:conf/acl/TangJLZ20}
Tang, H.; Ji, D.; Li, C.; and Zhou, Q. 2020.
\newblock Dependency Graph Enhanced Dual-transformer Structure for Aspect-based Sentiment Classification.
\newblock In \emph{Proceedings of the 58th Annual Meeting of the Association for Computational Linguistics, {ACL} 2020, Online, July 5-10, 2020}, 6578--6588. Association for Computational Linguistics.

\bibitem[{van~den Oord, Li, and Vinyals(2018)}]{DBLP:journals/corr/abs-1807-03748}
van~den Oord, A.; Li, Y.; and Vinyals, O. 2018.
\newblock Representation Learning with Contrastive Predictive Coding.
\newblock arXiv:1807.03748.

\bibitem[{Xiao et~al.(2024{\natexlab{a}})Xiao, Mao, Zhang, He, and Cambria}]{DBLP:conf/emnlp/Xiao0ZHC24}
Xiao, L.; Mao, R.; Zhang, X.; He, L.; and Cambria, E. 2024{\natexlab{a}}.
\newblock Vanessa: Visual Connotation and Aesthetic Attributes Understanding Network for Multimodal Aspect-based Sentiment Analysis.
\newblock In \emph{Findings of the Association for Computational Linguistics: {EMNLP} 2024, Miami, Florida, USA, November 12-16, 2024}, 11486--11500. Association for Computational Linguistics.

\bibitem[{Xiao et~al.(2024{\natexlab{b}})Xiao, Wu, Xu, Li, Jin, and He}]{DBLP:journals/inffus/XiaoWXLJH24}
Xiao, L.; Wu, X.; Xu, J.; Li, W.; Jin, C.; and He, L. 2024{\natexlab{b}}.
\newblock Atlantis: Aesthetic-oriented multiple granularities fusion network for joint multimodal aspect-based sentiment analysis.
\newblock \emph{Inf. Fusion}, 106: 102304.

\bibitem[{Yang, Na, and Yu(2022)}]{DBLP:journals/ipm/YangNY22}
Yang, L.; Na, J.; and Yu, J. 2022.
\newblock Cross-Modal Multitask Transformer for End-to-End Multimodal Aspect-Based Sentiment Analysis.
\newblock \emph{Inf. Process. Manag.}, 59(5): 103038.

\bibitem[{Yao et~al.(2024)Yao, Li, Ren, Wang, Liu, Sun, and Hou}]{DBLP:journals/corr/abs-2405-20985}
Yao, L.; Li, L.; Ren, S.; Wang, L.; Liu, Y.; Sun, X.; and Hou, L. 2024.
\newblock DeCo: Decoupling Token Compression from Semantic Abstraction in Multimodal Large Language Models.
\newblock arXiv:2405.20985.

\bibitem[{Yu, Chen, and Xia(2023)}]{DBLP:journals/taffco/YuCX23}
Yu, J.; Chen, K.; and Xia, R. 2023.
\newblock Hierarchical Interactive Multimodal Transformer for Aspect-Based Multimodal Sentiment Analysis.
\newblock \emph{{IEEE} Trans. Affect. Comput.}, 14(3): 1966--1978.

\bibitem[{Yu and Jiang(2019)}]{DBLP:conf/ijcai/Yu019}
Yu, J.; and Jiang, J. 2019.
\newblock Adapting {BERT} for Target-Oriented Multimodal Sentiment Classification.
\newblock In \emph{Proceedings of the Twenty-Eighth International Joint Conference on Artificial Intelligence, {IJCAI} 2019, Macao, China, August 10-16, 2019}, 5408--5414. ijcai.org.

\bibitem[{Zhao et~al.(2023)Zhao, Li, Wu, Ouyang, Zhang, and Dai}]{DBLP:conf/emnlp/ZhaoLWOZD23}
Zhao, F.; Li, C.; Wu, Z.; Ouyang, Y.; Zhang, J.; and Dai, X. 2023.
\newblock {M2DF:} Multi-grained Multi-curriculum Denoising Framework for Multimodal Aspect-based Sentiment Analysis.
\newblock In \emph{Proceedings of the 2023 Conference on Empirical Methods in Natural Language Processing, {EMNLP} 2023, Singapore, December 6-10, 2023}, 9057--9070. Association for Computational Linguistics.

\bibitem[{Zhou et~al.(2023)Zhou, Guo, Liu, Yu, Zhang, and Yuan}]{DBLP:conf/acl/ZhouGLYZY23}
Zhou, R.; Guo, W.; Liu, X.; Yu, S.; Zhang, Y.; and Yuan, X. 2023.
\newblock AoM: Detecting Aspect-oriented Information for Multimodal Aspect-Based Sentiment Analysis.
\newblock In \emph{Findings of the Association for Computational Linguistics: {ACL} 2023, Toronto, Canada, July 9-14, 2023}, 8184--8196. Association for Computational Linguistics.

\bibitem[{Zhu et~al.(2024)Zhu, Sun, Gao, Yi, and He}]{DBLP:conf/ijcai/ZhuSGY024}
Zhu, L.; Sun, H.; Gao, Q.; Yi, T.; and He, L. 2024.
\newblock Joint Multimodal Aspect Sentiment Analysis with Aspect Enhancement and Syntactic Adaptive Learning.
\newblock In \emph{Proceedings of the Thirty-Third International Joint Conference on Artificial Intelligence, {IJCAI} 2024, Jeju, South Korea, August 3-9, 2024}, 6678--6686. ijcai.org.

\end{thebibliography}

\end{document}